\documentclass[10pt,twocolumn,letterpaper]{article}
\pdfoutput=1
\usepackage{cvpr}
\usepackage{times}
\usepackage{epsfig}
\usepackage{graphicx}
\usepackage{amsmath}
\usepackage{amssymb}

\usepackage[pagebackref=true,breaklinks=true,letterpaper=true,colorlinks,bookmarks=false]{hyperref}

\cvprfinalcopy % *** Uncomment this line for the final submission

\def\cvprPaperID{2150} % *** Enter the CVPR Paper ID here

\ifcvprfinal\fi
\begin{document}

%%%%%%%%% TITLE
\title{A Frequency Domain Neural Network for Fast Image Super-resolution}

\author{Junxuan Li${^1}$\thanks{Corresponding author (email: u5990546@anu.edu.au)}\ \ \ \ \ \ \ \ \ \ \ \ \ Shaodi You$^{1,2}$
\ \ \ \ \ \ \ \ \ \ \ \ \ Antonio Robles-Kelly$^{1,2}$\\
% \textsuperscript{1}Australian National University\\
% {\tt\small u5990546@anu.edu.au}
% \and
% \textsuperscript{2}Data61, CSIRO\\
% {\tt\small \{shaodi.you, antonio.robles-kelly\}@data61.csiro.au}
$^1$College of Eng. and Comp. Sci., Australian National University, Canberra, ACT 2601, Australia\\
$^2$Datat61-CSIRO, Black Mountain Laboratories, Acton, ACT 2601, Australia\\
}

\maketitle
%\thispagestyle{empty}

%%%%%%%%% ABSTRACT
\begin{abstract}
   In this paper, we present a frequency domain neural network for image super-resolution. The network employs the convolution theorem so as to cast convolutions in the spatial domain as products in the frequency domain. Moreover, the non-linearity in deep nets, often achieved by a rectifier unit, is here cast as a convolution in the frequency domain. This not only yields a network which is very computationally efficient at testing but also one whose parameters can all be learnt accordingly. The network can be trained using back propagation and is devoid of complex numbers due to the use of the Hartley transform as an alternative to the Fourier transform. Moreover, the network is potentially applicable to other problems elsewhere in computer vision and image processing which are often cast in the frequency domain. We show results on super-resolution and compare against alternatives elsewhere in the literature. In our experiments, our network is one to two orders of magnitude faster than the alternatives with an imperceptible loss of performance.
\vspace{-0.8cm}
\end{abstract}

%%%%%%%%% BODY TEXT
\section{Introduction}

Image super-resolution is a classical problem which has found 
application in areas such as video processing \cite{eren:97}, light field imaging \cite{bishop:09}
and image reconstruction \cite{farsiu:2003}. 
%\shaodi{How do with want to pinch the sales point? Image super-res of frequent domain deep learning? }

Given its importance, super-resolution has attracted ample attention in the image processing and
computer vision community. Early approaches to super-resolution are 
often based upon the rationale that higher-resolution images have a frequency domain representation whose
higher-order components are greater than their lower-resolution analogues. Thus, methods such as that in \cite{tsai:84} exploited the shift and aliasing properties of the Fourier transform to recover a super-resolved
image. Kim {\it et al.} \cite{kim:90} extended the method in  \cite{tsai:84} to settings where noise and spatial blurring are present in the input image. In a related development, in \cite{bose:93}, super-resolution in the frequency domain is effected using Tikhonov regularization. 

Alternative approaches, however, effect super-resolution by {\it aggregating} multiple frames with complementary spatial information or by relating the higher-resolved image to the lower resolution one by a sparse linear system. For instance, Baker and Kanade \cite{baker:2002} formulated the problem in a regularization setting where the examples are constructed using a pyramid approach. Protter {\it et al.} \cite{protter:2009} used block matching to estimate a motion model and use exemplars to recover super-resolved videos. Yang {\it et al.} \cite{yang2010image} used sparse coding to perform super-resolution by learning a dictionary that can then be used to produce the output image, by linearly combining learned exemplars.

Note that, the idea of super-solution ``by example'' can be viewed as hinging on the idea of learning functions so as to map a lower-resolution image to a higher-resolved one using exemplar pairs. This is right at the center of the philosophy driving deep convolutional networks, where the net is often considered to learn a non-linear mapping between the input and the output. In fact, Dong {\it et al.} present in \cite{dong2016image} a deep convolutional network for single-image super-resolution which is equivalent to the sparse coding approach in \cite{yang2010image}. In a similar development, Kim {\it et al.} \cite{kim2016accurate} present a deep convolutional network inspired by VGG-net \cite{simonyan:2014}. The network in \cite{kim2016accurate} is comprised of 20 layers so as to exploit the image context across image regions. In \cite{kappeler:2016}, a multi-frame deep network for video super-resolution is presented. The network employs motion compensated frames as input and single-image pre-training. 

\begin{figure*} % this figure should the frequency selector map configuration
\includegraphics[width = \textwidth]{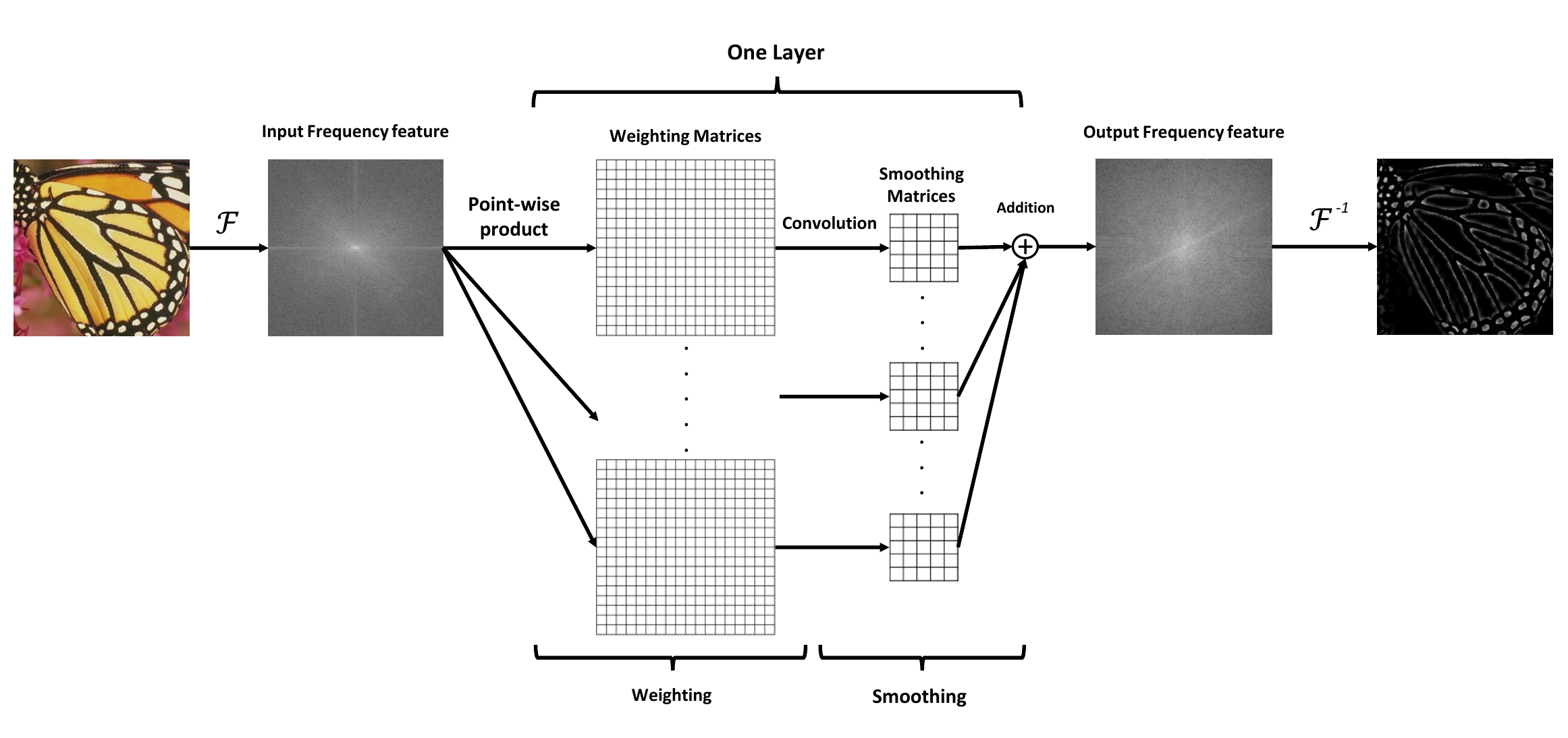}
\caption{Diagram of our network with a single-layer. The input feature map obtained by transforming the image into the frequency domain is point-wise multiplied with the weighting matrices. After the weighting operation, the result is convolved  and added to obtain the output frequency feature map that can then be transformed into the spatial domain to recover the predicted residual. Note that, in our network, the weighting  matrices are the same size as the input feature map. }
\label{fig:layer}
\vspace{-0.3cm}
\end{figure*}

\section{Contribution}

Here, we present a computationally efficient frequency domain deep network for super-resolution which, at input, takes the frequency domain representation of the lower resolution image and, at output, recovers the residual in the frequency domain. This residual can then be added to the lower resolution input image to obtain the super-resolved image. The network presented here is somewhat related to those above, but there are a number of important differences with respect to other approaches. Its important to note that:
\begin{itemize}
\item Following our frequency domain interpretation of deep networks, the convolutions in other networks are treated here as multiplications. This has the well known advantages of lower computational cost and added computational efficiency. 
\item Following the frequency domain treatment to the problem as presented here, the non-linearity in the network is given by convolutions in the frequency domain. This contrasts with the work in \cite{rippel2015spectral}, which employs spectral pooling instead.
\item In contrast with deep network architectures elsewhere, where the non-linearity is often attained using an activation function such as a rectified linear unit (ReLU) \cite{glorot2011deep}, we can learn the analogue convolutional parameters in the frequency domain in a manner akin to that used by CNNs in the spatial domain.
\item We use residual training since its not only known to deal with the vanishing gradients well and often improve convergence, but its also particularly well suited to our net. This is following the notion that the lower resolved image lacks the higher frequencies in high-resolution imagery and, thus, these can be learned by the network based on the residual.  
\item Finally, we employ the Hartley transform as an alternative to the Fourier transform so as to avoid the need to process imaginary numbers.% without any loss of generality. %Thus, in our implementation, we use the Hartley transform without any loss of generality. This is since the Hartley transform is closely related to the Fourier transform and does have an analogue convolution theorem. 
\end{itemize}

%\section{Related Work }

\begin{figure*} % this figure should the frequency selector map 
\centering
\includegraphics[width = \textwidth]{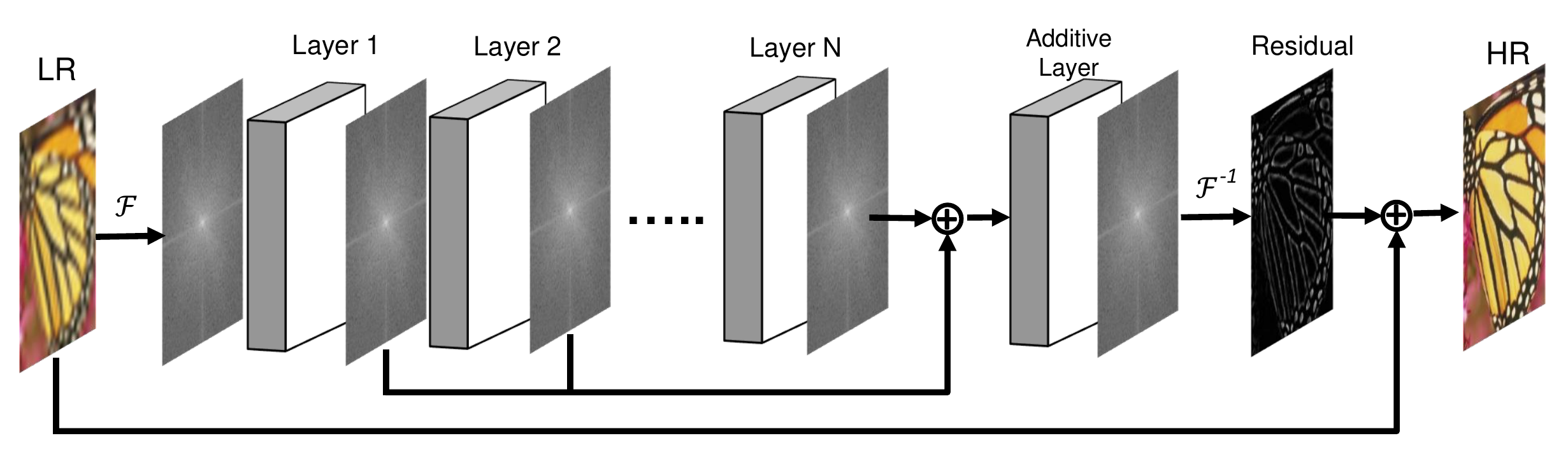}
\caption{Simplified diagram of our network with multiple layers. Each of these layers accounts for a product-convolution-addition step as presented in Figure \ref{fig:layer}. Here, the lower-resolved image is transformed into the frequency domain and then, after the final additive layer, the predicted residual is transformed back into the spatial domain. The residual and input images are then added to obtain the higher-resolved output image.}
\label{fig:network}
\vspace{-0.3cm}
\end{figure*}

\section{Spatial and Frequency Domains}
\subsection{Convolutional Neural Networks}
Note that most of the convolutional neural networks nowadays are variants of ImageNet \cite{ImageNetAlex2012}. Moreover, from a signal processing point of view, these networks can be considered to work on the ``spatial'' image domain\footnote{Here, we adopt the terminology often used in image processing and computer vision when comparing spatial and frequency domain representations. We have done this so as to be consistent with longstanding work on integral transforms such as Fourier, Cosine and Mellin transforms elsewhere in the literature.}. In these networks, each layer is comprised of a set of convolutional operations followed by an activation function.

To better understand the relationship between these networks in the spatial domain and ours, which operates in the frequency domain, recall that the two-dimensional discrete convolutions at the $i^{th}$ layer can be expressed as 
\begin{equation}
(f*g_j)[u,v] = \sum_{m=-M}^{M}\sum_{n=-N}^{N}  f[m,n] g_j[u-m,v-n]  
\label{conv_eq}
\end{equation}
where $f$ denotes the feature map delivered by the previous layer, \textit{i.e.} that indexed $i-1$, in the network and $g_j$ is the convolutional kernel of order $(2M+1)\times(2N+1)$ in the corresponding layer. In the equation above, we have used $u$ and $v$ as the spatial coordinates (rows and columns) in the image under consideration. 

At each layer, this convolutional stage is followed by an activation function which induces the non-linearity in the behavior of the network. Nonetheless there are a number of choices of activation function, \textit{i.e.} sigmoid, binary, identity, etc., the most widely used one is  ReLU, which can be expressed mathematically using a product with a Heaviside function as follows 
\begin{eqnarray}
\label{relu_eq}
\text{ReLU}[u,v]&=&\text{max}(0,(f*g_j)[u,v])\\
\nonumber
&=&(f*g_j)[u,v] \ \text{HS}((f*g_j)[u,v]) 
\end{eqnarray}
where $\text{HS}(x)$ is the Heaviside function which yields $1$ if $x>0$, $0.5$ if $x=0$  and $0$ if $x<0$. In the expressions above, and so as to be consistent with Equation \ref{conv_eq}, we have used $(f*g_j)[u,v]$ as the input to the rectifier unit.

\subsection{Fourier Transform}
Equations \ref{conv_eq} and \ref{relu_eq} hint at the use of the convolution theorem of the Fourier transform to obtain an analogue of spatial domain CNNs in the frequency domain. Further, the convolution theorem as widely known for the Fourier transform has similar instantiations for other closely related integral transforms. For now, and for the sake of clarity, we will focus on the Fourier transform. Later on in the paper, we will elaborate further on the use of the Hartley transform as an alternative to the Fourier transform in our implementation. 

% This is since the function $f(m,n)$ be written as Eq~\ref{four_t} and discrete form in Eq~\ref{conv_dis_eq}. Where $\mathcal{F}\{\cdot\}$ is usually known as the  frequency spectrum function and is a complex function. Its modulus $|\mathcal{F}\{\cdot\}|$ represent the  amplitude  of each frequency in the signal. 
% \begin{equation}
% %\mathcal{F}\{f(x,y)\} = \iint_{-\infty}^{\infty} f(x,y) e^{-j2\pi(ux + vy)} dxdy \label{four_t}\\
% \mathcal{F}\{f(m,n)\} = \sum_{m=-\infty}^{\infty}\sum_{n=-\infty}^{\infty} f[m,n] e^{-j2\pi(um+vn)} 
% \label{conv_dis_eq}
% \end{equation}

% \subsection{Convolution Theorem}
The convolution theorem states that given two functions in spatial (time) domain, their convolution is given by their point-wise multiplication in the frequency domain. For the sake of consistency with Equations \ref{conv_eq} and \ref{relu_eq}, let $f$ and $g_j$ be the two spatial domain functions under consideration and denote the Fourier transform operator as $\mathcal{F}$, then we have
\begin{equation}
\mathcal{F}\{ f * g_j \} = \mathcal{F}\{ f\} \cdot \mathcal{F}\{ g_j\} 
\label{conv_prod}
\end{equation}
where $\cdot$ denotes point-wise product.

Moreover, the converse relation also holds, whereby a product in the spatial domain becomes a convolution in the frequency domain. \textit{i.e.} 
\begin{eqnarray}
\label{prod_conv} 
\mathcal{F}\{  (f * g_j) \cdot h_j \} &=& \mathcal{F}\{ f * g_j\} * \mathcal{F}\{ h_j\}\\
\nonumber
&=&\big(\mathcal{F}\{ f\} \cdot \mathcal{F}\{ g_j\}\big)* \mathcal{F}\{ h_j\}
\end{eqnarray}
where we have   
%used the shorthand $\mathcal{Q} =\mathcal{F}\{ f * g_j\}$ and 
employed $h_j$ to denote a generic activation function which, in our case can also be learnt. Note that the second line in the equation above follows from substituting Equation \ref{conv_prod} into the first line.

%\subsection{Volterra series}
%Volterra series\cite{barrett1977bibliography}

\section{Network Structure}

Note that, in the Equation \ref{prod_conv}, the term $\mathcal{F}\{ g_j\}$ acts as a ``weighting'' operation in the frequency domain. That is, through the point-wise product operation, it curtails or amplifies the frequency domain components of $\mathcal{F}\{ f\}$. These frequency weighting operations take  place of the original convolution operation in spatial domain convolutional neural networks such as ImageNet \cite{ImageNetAlex2012}. Similarly, the nonlinear operator given by the rectifier units in CNNs is now substituted, in the frequency domain, by a convolution. This can be viewed as a smoothing or regularization operation in the frequency domain.

In Figure \ref{fig:layer}, we show a single-layer instantiation of our frequency domain network. Note that, at input, the image is transformed into the frequency domain. Once this has been effected, the frequency weighting step takes place, \textit{i.e.} the pointwise multiplication operation, and then a convolution is applied. Once the outputs of all the convolutional outputs are obtained, they are summed together by an additive layer and the inverse of the frequency domain transform is applied so as to obtain the final result. Its worth noting that we have not incorporated a spectral or spatial pooling layer in our network. This is not a major issue in our approach since these pooling layers are often used for classification tasks \cite{ImageNetAlex2012} whereas in other applications, such as super-resolution, pooling is seldom used \cite{dong2016image}.

%A prediction layer take place the full connective layer. We will go through the detail of these new layers in this section.
\subsection{Weighting}

As mentioned above, the product $\mathcal{Q} =\mathcal{F}\{ f\}\cdot \mathcal{F}\{g_j\}$ can be viewed as a frequency weighting operation equivalent to the convolution operation in time domain. As before, consider a feature map $f$ at a given layer in the network and the $j^{th}$ convolutional kernel $g_j$. 

For the layer under consideration, the  product $\mathcal{Q}$ will take the frequency domain of the feature map $\mathcal{F}\{ f\}$ as an input and point-wise multiply it by a wight matrix given by the values of $\mathcal{F}\{g_j\}$. In practice, both, $\mathcal{F}\{ f\}$ and $\mathcal{F}\{g_j\}$ can be viewed as matrices which are the same size. This is important since it permits us to pose the problem of performing the forward pass and backpropagation steps in a manner analogous to that used in CNNs operating in the time domain. 

To see this more clearly, denote as $\mathbf{F}^i$ the input matrix corresponding to $\mathcal{F}\{ f\}$ to the $i$th layer of our frequency domain network. Similarly, let the $j^{th}$ weight matrix corresponding to the coefficients of  $\mathcal{F}\{g_j\}$ be $\mathbf{W}_j$. The output of the product of the two matrices is another matrix, which we denote $\mathbf{Q}$ and whose entries indexed $l,k$ are given given by
\begin{equation}
Q(l,k) = F^i(l,k) W_j(l,k)  + B_j(l,k)
\label{eq:fre_sel}
\end{equation}
where $F^i(l,k)$ and $W_j(l,k)$ are the entries indexed $l,k$ of the matrices $\mathbf{F}^i$ and $\mathbf{G}_j$, respectively, and we have introduced the bias matrix $\mathbf{B}_j$ with entries $B_j(l,k)$.

Moreover, the Fourier transform of an image, being real and non-negative, is conjugate-symmetric\footnote{It can be shown in a straightforward manner that this symmetry property also applies to the Hartley transform.}. This is important since, by noting that $\mathbf{W}_j$ should be Hermitian, we can reduce the number of learnt weights by half. 

%By using the conjugate-symmetric property of real-valued Fourier transforms we can write Equation \ref{eq:fre_sel} as follows
% \begin{equation}
% Q(l,k) = F^i(l,k) W_j(l,k)  + b_j
% \label{eq:fre_sel_sym}
% \end{equation}
% %w_{ij}[x-c_x,y-c_y] = w_{ij}[c_x-x,c_y-y]
% %\end{align}
% where $x,y$ denote the coordinate of weights in $w_{ij}$, $c_x,c_y$ denote the coordinate of central point which is usually half length of corresponding axis.

% Hence the number of independent parameters in each selector decrease to half, which will speed up the convergence of network.

\subsection{Smoothing}

As shown in Figure \ref{fig:layer}, once the weighting operation is effected, a convolution in the frequency domain, analogous to the rectification operation in the spatial domain is applied. This is inspired upon Equation \ref{prod_conv}, which hints at the notion that we can express the ReLU as a product between a Heaviside function and its argument. Again, in practice, this can be expressed as follows
\begin{align}
R_j(l,k) = \sum_{m=-M}^M\sum_{n=-N}^N Q(l-m,k-n)C_j(m,n)
\label{eq:ournonlinear}
\end{align}
Where  $Q(l-m,k-n)$ is the corresponding entry of the matrix $\mathbf{Q}$ as presented in the previous section and $C_j(m,n)$ are the coefficients of the matrix $\mathbf{C}$ containing the values of $\mathcal{F}\{ h_j\}$.

From the equation above is straightforward to note that the entries of the matrix $\mathbf{R}$ are a linear combination of the values of $\mathbf{Q}$ where the matrix $\mathbf{C}$ can be viewed as a kernel that can be learnt. Thus, here, we consider the entries $C_j(m,n)$ of $\mathbf{C}_j$ as parameters that can be updated at each back-propagation step. This, in turn, allows us to learn both, the weights in $\mathbf{W}_j$ as well as the parameters in $\mathbf{C}_j$.

\begin{figure}[!t]
\centering
\includegraphics[width =0.45\textwidth]{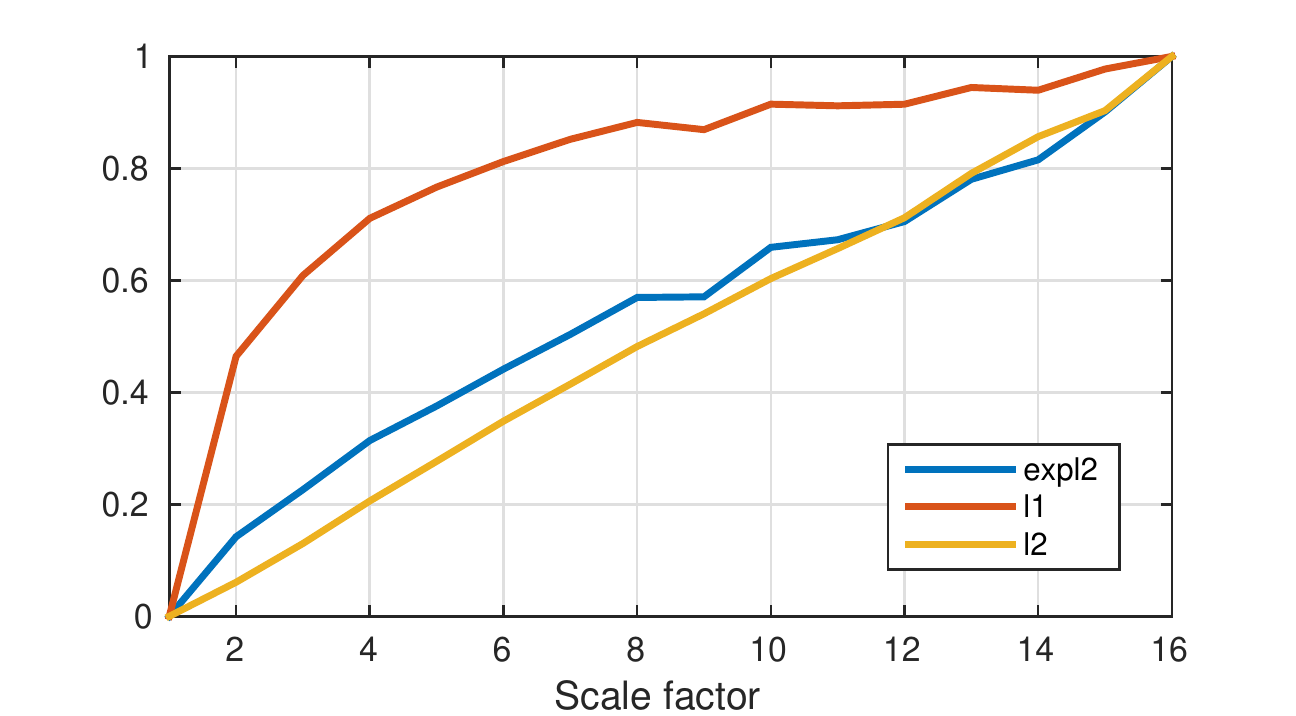}
\caption{Loss comparison as a function of low-to-high resolution factors. %The result is test on bicubic interpolation. 
All the loss function values, \textit{i.e.} $l_1$, $l_2$, Exp-$l_2$, are computed in frequency domain and normalized to unit at their extrema.}
\label{fig:losscomapre}
\vspace{-0.3cm}
\end{figure}

\begin{figure}[!b]
\includegraphics[width = 0.48\textwidth]{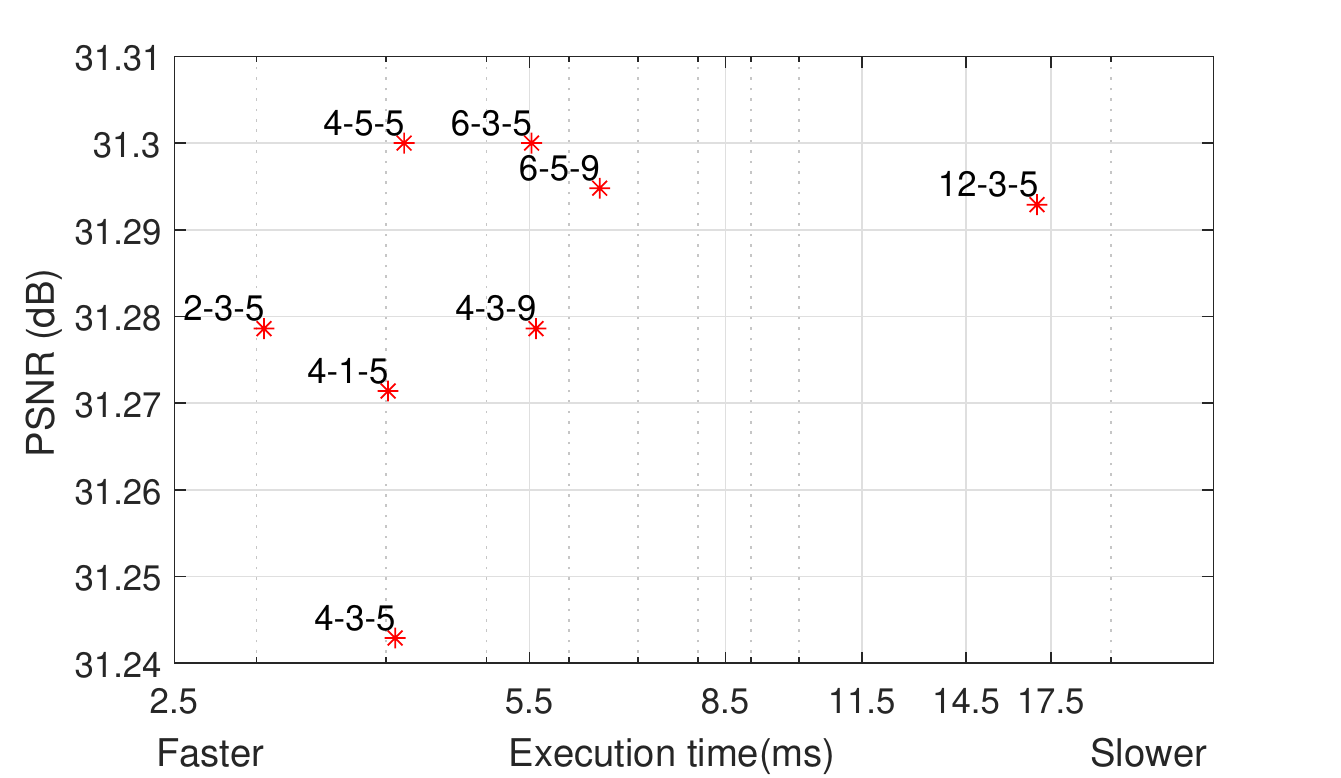}
\caption{Cross-validation performance of our network as a function of testing time. The text over a point denotes the number of layers $L$, smoothing matrix size $N$ and the number of weighting matrices per layer $K$, respectively. All the variants of the network were tested on the Set14 dataset with an upscaling factor of 2.}
\label{fig:exppars}
\end{figure}

\begin{figure*} 
\centering
\includegraphics[width = 0.9\textwidth]{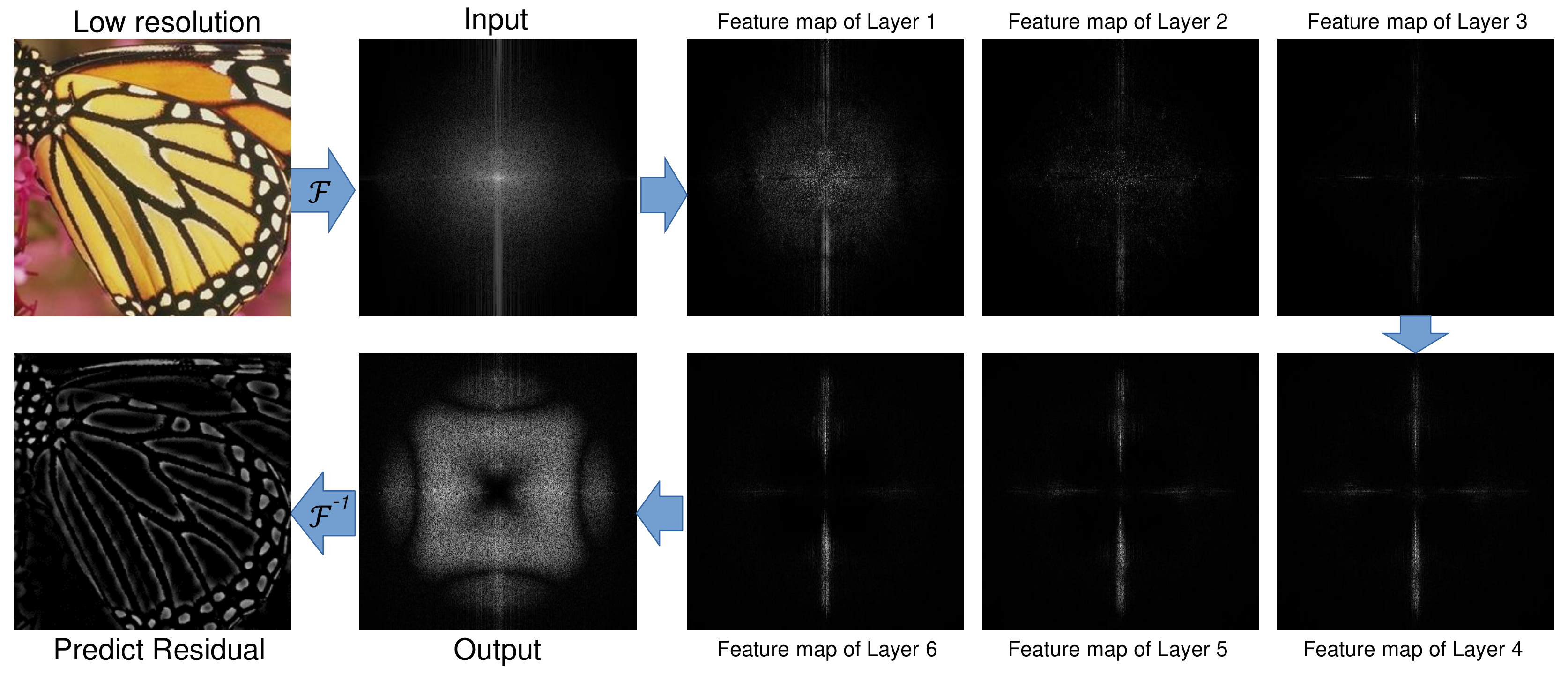}
\caption{Frequency domain feature maps for each layer in our network. Here we also show the input image and predicted residual in the spatial domain.}
\label{fig:outlayer}
\vspace{-0.3cm}
\end{figure*}

\subsection{Additive Layer}

Recall that, in applications such as super-resolution, frequency domain approaches aim at recovering or predicting a whole frequency map corresponding to either the enhanced or super-resolved image. As a result, instead of a prediction layer, our network adds the output of all the network feature maps into a single one at output and then applies a final frequency weighting operation.
This additive layer can be expressed as follows
\begin{equation} %  equation here to describe how to add feature map before together 
\mathbf{P} =  \bigg(\sum_{i=1}^L \alpha_i \mathbf{S}^i\bigg)\odot \mathbf{W}_L  
\label{eq:P}
\end{equation}
where $L$ is the number of layers in the network, $\odot$ denotes the Hadamard (entrywise) product, $\mathbf{W}_L$ is the final frequency weighting matrix, $\mathbf{P}$ is the prediction of our network in the frequency domain and $\alpha_i$ is the weight that controls the contribution of the $i^{th}$ layer feature map to the output. 

In the equation above, $\mathbf{S}^i$ is given by summing  over all the matrices $\mathbf{R}_j$ for each layer $i$. In our network, we do not require this sum to have independent weights since these can be absorbed, in a straightforward manner, into the matrices $\mathbf{C}_j$. Thus, for this layer, in practice, we only learn the matrix $\mathbf{W}_L$.% and the weights $\alpha_i$. 
%This, again, can be done using backpropagation.

In Figure~\ref{fig:network} we show the simplified diagram of our complete network. Note that the output of each layer is a frequency feature map which is the same size as the input. The output of each layer is then added and weighted by the additive layer. We then compute the spatial domain residual by applying the inverse transform to the frequency domain output of our network and add the predicted residual to the input image so as to compute the super-resolved output.

%\section{Experiments}
\section{Implementation and Discussion}
%We apply our network on the image super-resolution task. Since this task is  mainly to recover high frequency information from lower one. 

\subsection{Hartley vs Fourier Transform}
As mentioned earlier, the implementation of our network makes use of the Hartley transform \cite{HartleyTrans42} as an alternative to the Fourier transform. The reasons for this are twofold. Firstly, the Hartley transform is closely related to the Fourier transform and, hence, shares analogue properties. Secondly, the Hartley transform takes at input real numbers and delivers, at output, real numbers. This has the advantage that, by using the Hartley transform, we do not need to cater for complex numbers in our implementation. 

Here, we have used the fast Hartley transform introduced by Bracewell  \cite{bracewell:84}. The Hartley transform can be expressed using the real $\mathfrak{R}\{\cdot\}$ and imaginary $\mathfrak{I}\{\cdot\}$ parts  of the Fourier transform as follows
\begin{equation}
\mathcal{H}\{f\} = \mathfrak{R}\{\mathcal{F}\{f\}\} - \mathfrak{I}\{\mathcal{F}\{f\}\} 
\label{hartleytrans}
%f &= \mathcal{H} \{\mathcal{H}\{f\}\}
\end{equation}
where we have used $f$ for the sake of consistency with previous sections. Moreover, the Hartley transform is an involution, that is, the inverse is given by itself, \textit{i.e.}   
%\begin{equation}
%\mathcal{H}\{f\} &= \mathfrak{R}\{\mathcal{F}\{f\}\} - \mathfrak{I}\{\mathcal{F}\{f\}\} \label{hartleytrans}\\
$f = \mathcal{H} \{\mathcal{H}\{f\}\}$.  
%\end{equation}

It is worth noting that, from Equation \ref{hartleytrans}, its straightforward to show that, since the Fourier transform is linear as are the matrices $\mathbf{W}_j$, the weighting operation in our network applies, in a straightforward manner to the Hartley transform without any loss of generality. In the case of the Hartley transform, the convolution theorem has the same form as that of the Fourier transform  \cite{bracewell:84}, and, hence, we can write Equation \ref{prod_conv} as follows
\begin{equation}
\label{prod_conv_hartley} 
%\mathcal{F}\{  (f * g_j) \cdot h_j \} &=& \mathcal{F}\{ f * g_j\} * \mathcal{F}\{ h_j\}\\
%\nonumber
\mathcal{H}\{  (f * g_j) \cdot h_j \}=\big(\mathcal{H}\{ f\} \cdot \mathcal{H}\{ g_j\}\big)* \mathcal{H}\{ h_j\}
\end{equation}

% describe in Eq~\ref{hartleytrans}. Where $\mathcal{H}\{f\}$ denote the Hartley transform and $\mathfrak{R}\{x\}, \mathfrak{I}\{x\}$ denote  the real part and image part of a complex number $x$. The inverse of Hartley transform only need to apply another Hartley transform to itself.

\begin{figure}[!t]
\centering
\includegraphics[width = \columnwidth]{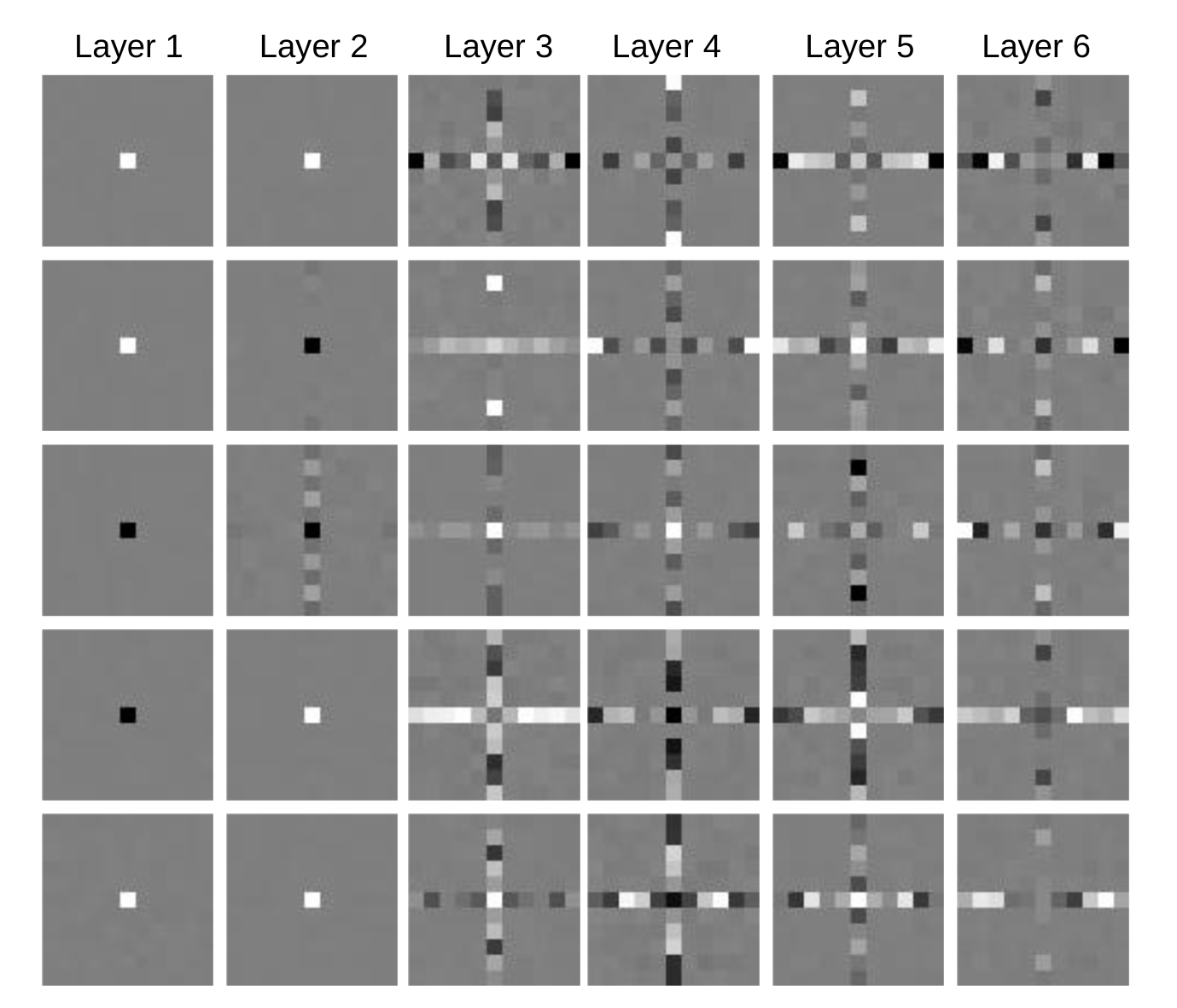}
\caption{Smoothing matrices across the 6 layers in our network trained on the full dataset with an upscaling factor of 2. In the figure, the columns account for different layers. }
\label{fig:nonlinearkernel}
\vspace{-0.3cm}
\end{figure}

\begin{figure*}
\centering
\includegraphics[width = 0.8\textwidth,height = 0.72\textheight]{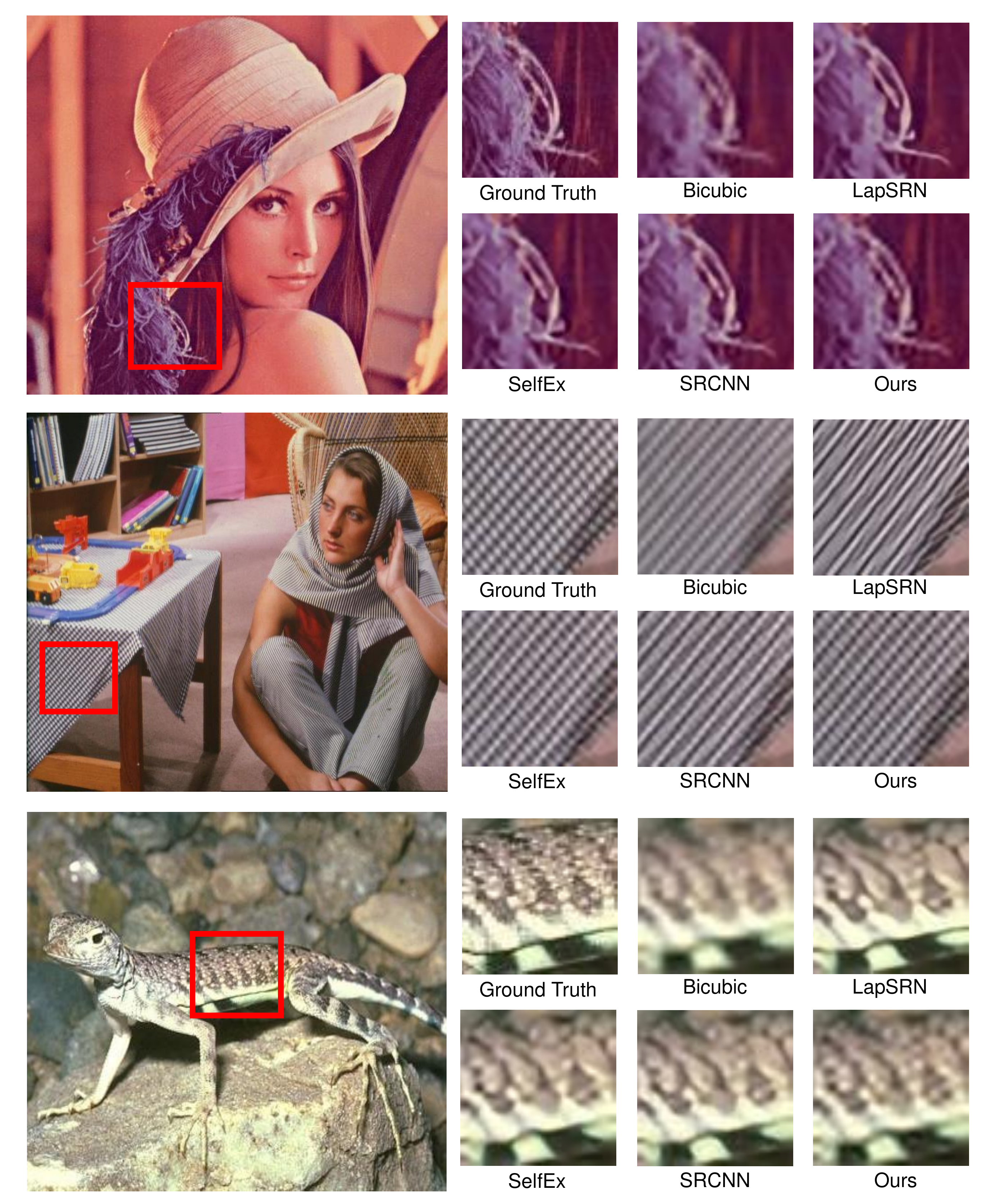}
\caption{Qualitative comparison for an upscaling factor of $3$ on ``Lena'', ``Barbara'' and a sample image from the B100 dataset. Here we show a detail of the results yielded by our method and those delivered by LapSRN \cite{LapSRN}, SelfEx \cite{huang2015single}, SRCNN \cite{dong2016image}. %We also show the high-resolution image ground truth. %Exemplar based methods tend to over exaggerate edge and leads to error.
}
\label{fig:compare}
\vspace{-0.3cm}
\end{figure*}

\subsection{Training}

For training our network, we have used the mini-batch gradient descent method proposed by LeCun \etal \cite{lecun1998gradient}. When training our 
network, the layers at the end, \textit{i.e.} those closer to the final additive layer, tend to have a gradient that is small in magnitude as compared to the first layers.  The reason being that, due to the architecture of our network, as shown in Figure \ref{fig:network}, the first layers (those closer to the input) will accumulate the gradient contributions of the deeper layers in the network. This is a problem akin to that in \cite{dong2016image}, which was tackled by the VDSR \cite{kim2016accurate} approach  by applying gradient clipping in the back propagation step \cite{pascanu2013difficulty}. As a result, we follow  \cite{kim2016accurate} and normalize the gradient at each layer to a fixed range $(-\theta,\theta)$. Thus, the parameters at the layer can only change within a fixed range $(-\gamma\theta,\gamma\theta)$. In our implementation, we set $\theta = 10^{3}$ and $\gamma=10^{-5}$ for all layers. %For training, we do not apply batch normalisation. 

\subsection{Residual and Choice of Loss Function}

For our loss function, we consider a number of alternatives. These are the L1 ($l_1$), L2 ($l_2$) and L2 with exponential decay ($\text{Exp-}l_2$) loss functions. 
%Let $x$ denote a interpolated low-resolution image in frequency domain and $y$ denote a high-resolution image in frequency domain. We aim to learn our model $f$ that predict $f(x)$ which is the estimate of HR image in frequency domain.
%The equality of image recovery in frequency domain is never been defined before. We have tried several loss functions in frequency domain like $l_1$, $l_2$,
These are defined as follows
\begin{align}
l_1 &\stackrel{ \text{def}}{=} || \mathbf{I} - \mathbf{I}^* ||_1   \label{eq:l1}\\
l_2 &\stackrel{ \text{def}}{=} || \mathbf{I} - \mathbf{I}^* ||_2^2 \label{eq:l2}\\
\text{Exp-}l_2 &\stackrel{ \text{def}}{=} e^{\beta \omega_x} || \mathbf{I} - \mathbf{I}^* ||_2^2\label{eq:expl2}
\end{align}
where $||\cdot||_p$ denotes the $p$-norm under consideration, $\beta$ is a hyper-parameter, $\mathbf{I}$ is the matrix corresponding to the ground-truth high resolution image in the frequency domain and $\mathbf{I}^*$ accounts for the frequency domain image recovered by our network. This is yielded by the sum of the prediction $\mathbf{P}$ of our network as given in Equation \ref{eq:P} and the input to our network in the frequency domain, \textit{i.e.} $\mathbf{F}^1$, which can be expressed as %follows
\begin{equation}
%$
\mathbf{I}^* = \mathbf{P} + \mathbf{F}^1%$.   
\end{equation}

It is worth mentioning in passing that, in accordance with the observations made in \cite{kim2016accurate}, we also find that the use of residual learning improved the accuracy and converge speed of our network in the frequency domain. In Figure~\ref{fig:losscomapre}, we show a comparison of the three loss functions under consideration. In the figure, we show the loss values, normalized to unity, as a function of the low-resolution image scale factor of $ \mathbf{I}$ with respect to $\mathbf{I}$. In the figure, we have set $\beta=0.01$. Note that the L2 loss is almost linear with respect to the scale factor. Moreover, in our experiments, we found that the L2 loss  performed the best with respect to both, convergence and speed. As a result, all the experiments shown hereafter employ the L2 loss.

\begin{table*}
\begin{center}
{\small
\begin{tabular}{|l|c|c|c|c|c|c|c|}
\hline
  & & Bicubic & A+\cite{timofte2014a} & SRCNN\cite{dong2016image} & VDSR\cite{kim2016accurate} & LapSRN\cite{LapSRN} & Ours\\
Dataset & Scale & PSNR/SSIM & PSNR/SSIM & PSNR/SSIM  & PSNR/SSIM  & PSNR/SSIM  & PSNR/SSIM\\
\hline\hline
  		& 2	& 33.66/0.930 & 36.54/0.954 & 36.66/0.954 & 37.53/0.958 & 37.25/0.957 & 35.20/0.943   \\
Set5\cite{bevilacqua2012low} 	& 3 & 30.39/0.868 & 32.58/0.908 & 32.75/0.909 & 33.66/0.921 & 34.06/0.924 & 31.42/0.883\\
  		& 4 & 28.42/0.810  & 30.28/0.860 & 30.48/0.862 & 31.35/0.883 & 31.33/0.881 & 29.35/0.827\\
\hline\hline
  		& 2 & 30.24/0.868 & 32.28/0.905 & 32.42/0.906 & 33.03/0.912 & 32.96/0.910 & 31.40/0.895\\
Set14\cite{zeyde2010single} 	& 3 & 27.55/0.774 & 29.13/0.818  & 29.28/0.820 & 29.77/0.831 & 29.97/0.836 & 28.32/0.802\\
  		& 4 & 26.00/0.702 & 27.32/0.749 & 27.49/0.750  & 28.01/0.767 & 28.06/0.768  & 26.62/0.727\\
\hline\hline
  		& 2 & 29.56/0.843 & 31.21/0.886 & 31.36/0.887 & 31.90/0.896 &  31.68/0.892 & 30.58/0.877\\
B100\cite{MartinFTM01} & 3 & 27.21/0.738 & 28.29/0.783  & 28.41/0.786  & 28.82/0.797 & 28.92/0.802 & 27.79/0.772\\
  		& 4 & 25.96/0.667 & 26.82/0.708   & 26.90/0.710  & 27.29/0.725 & 27.22/0.724 & 26.42/0.696  \\
\hline
\end{tabular}}
\end{center}
\caption{Quantitative evaluation of our method as compared to state-of-art super-resolution algorithms. Here we show the average PSNR/SSIM for upscale factors of  $2$, $3$ and $4$ for each of the three testing datasets. }
\label{tab:performance}
\vspace{-0.2cm}
\end{table*}

\section{Experiments}

\subsection{Datasets}

Recall that DRRN\cite{taiimage} and VSDN\cite{kim2016accurate} use 200 images in Berkeley Segmentation Dataset\cite{MartinFTM01} combined with 91 images from Yang \etal \cite{yang2010image}. This set of images has also been used for training in other approaches \cite{LapSRN,schulter2015fast}. These methods often use techniques such as data augmentation, \textit{i.e.} application of transformations such as rotation, scaling and flipping transformations to generate novel images so as to complement the ones in the dataset. 
It is important to note, however, that these rotation, scaling and flipping in the spatial domain become frequency shift and scaling operations. Moreover, the dataset above, comprised of 291 images and their augmentation is, in practice, too small to allow for cross-validation of parameters.

Thus, here we have opted for a two-stage process using 5000 randomly selected images from the Pascal VOC2012 dataset \cite{Everingham10} for training and Set5 \cite{bevilacqua2012low},   Set14 \cite{zeyde2010single} and B100 \cite{MartinFTM01} for testing. The first stage corresponds to a cross-validation process of the parameters used in our network employing 800 images out of the 5000 in our dataset for training and Set14 for testing. Also, for cross-validation, we have resized the image to $360\times 480$ and used a scale factor of 2 on the dataset so as to obtain the images that are used as input to our network. After cross-validation, and once the parameters have been selected, the second stage is to proceed to train and test our network on the whole dataset and the three testing sets. In all our experiments, we have taken the color imagery and performed super-resolution on the \textbf{Y} channel in the \textbf{YCbCr} space \cite{poynton2012digital}. 
% For training other scale factors like $3\times, 4\times$ super-resolution, we resize image to $360\times 480$.

\subsection{Parameter Selection}

We have selected, through cross-validation, the number of layers $L$ in the network, the size of the matrices $\mathbf{C}_j$ and the number of weighting matrices $K$ per layer. In all our experiments we have used squared matrices $\mathbf{C}_j$,\textit{i.e.} $N=M$, and chosen a base line network with $L = 4$, $N=3$, $K=5$. We have then progressively increased $L$, $N$ and $K$ so as to explore the trade-off between timing and performance. 

In Figure \ref{fig:exppars}, we show the PSNR as a function of timing for the combinations of $L$, $K$ and $N$ used in our cross-validation exercise. For the sake of clarity, the time axis is shown in a logarithmic scale. Note that, in general, the networks with $4$ or $6$ layers seem to deliver the best trade-off between performance and timing. In the figure, $L=4$, $N=5$ and $K=5$ performs the best while $L=6$, $N=2$ and $K=5$ also performs well. Thus, and bearing in mind that a deeper network is expected to perform better in a larger dataset, for all the experiments shown here onwards, we set $L=6$, $N=5$ and $K=5$. 

%The choosing of non-linear kernel size $k$ can be easy, since almost all experiments present the consistence in  high PSNR with larger non-linear kernel size. So  we set $k=11$ in our final model. But it does not  show much relationship between number of selector $s$ and performance so we just take $s=5$ as our setting.

\subsection{Network Behavior}
%Figure~\ref{fig:nonlinearkernel} shows all the smoothing matrices  

Now, we turn our attention to the behavior of the network in terms of the weighting and smoothing matrices. 
In Figure~\ref{fig:outlayer}, we show the feature maps in the frequency domain as yielded by each of the network layers, \textit{i.e.} the matrices $\mathbf{F}^i$. From the figure, we can appreciate that the feature map for the first layer mainly contains the low frequency information corresponding to the input image. As the layers go deeper, the feature maps become dominated by the high frequency components. This is expected since the aim of prediction in our net is given by the residual, which, in frequency domain, is mainly comprised by higher order frequencies. 

%with some lost in information alongside a `ring' shown in the figure. We belief   that the lost part is actually  what a bicubic interpolation  will lost in doing downsampling and upsampling. So those `noise' in frequency domain would be abandoned by our network.

\begin{figure*}
\centering
\includegraphics[width=\textwidth ]{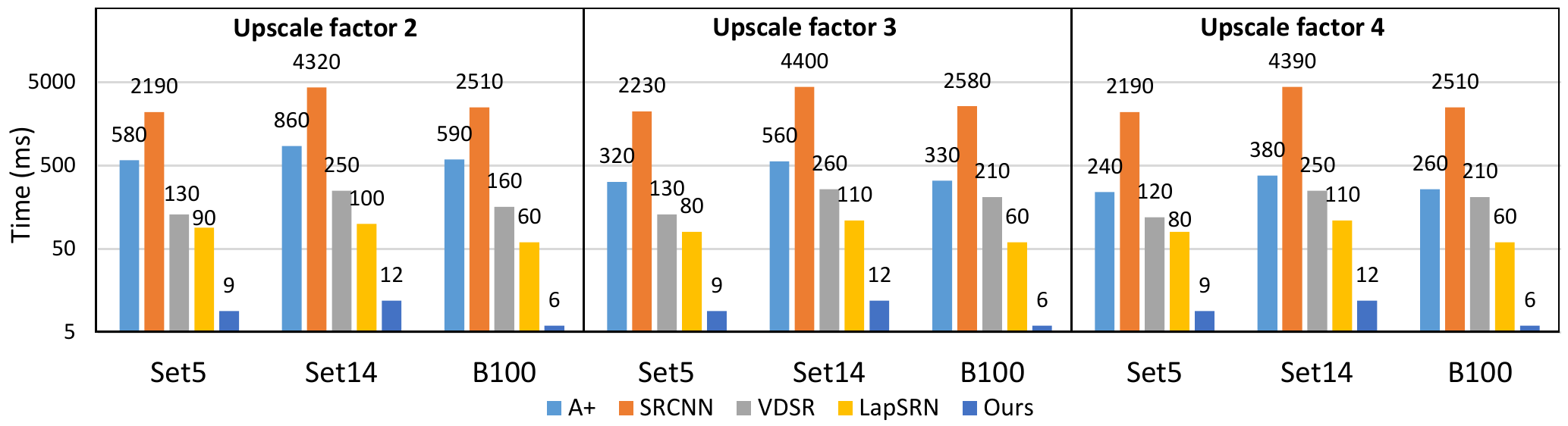}
\caption{Average per-image testing time for our method and the alternatives in milliseconds for upscale factors $2$, $3$ and $4$ on the three testing datasets. Our methods is a least 10 times faster than LapSRN, 20 times faster than VDSR, 40 times faster and A+ and 200 times faster than SRCNN.}%. (SRCNN uses the public slower implementation in CPU). }
\label{fig:timeplot}
\vspace{-0.3cm}
\end{figure*}

In Figure~\ref{fig:nonlinearkernel}, we show all the smoothing matrices in our network  
%with parameters  $L=6$, $N=5$, $K=5$ and an up-scaling factor 2 
after the training has been completed. Surprisingly, note that matrices in each layer all appear to behave slightly different. In the first layer, the matrices are very much a delta in the frequency domain, while as the layer index increases, they develop non-null entries mainly along the central column and row. This follows the intuition that, for the first layer, the convolution would behave as a multiplicative identity removing lower-order frequency components, whereas, for further layers, the main contribution to its output is given by the central rows and columns.

\subsection{Results}

Finally, we present our results on image super-resolution. To this end, we first show some qualitative results and then provide a quantitative evaluation of our network performance and testing timing.

In Figure \ref{fig:compare}, we show a detail of the results yielded by our network and a number of alternatives on the ``Barbara'', ``Lena'' images and a sample image from the B100 dataset. In all cases, we have applied an upscale factor of 3 and show, on the left-hand panel the full ground truth image. Note that our method yields results that are quite comparable to the alternatives. Moreover, for the detail of the ``Barbara'' image, the aliasing does not have a detrimental effect on the results. This contrasts with LapSRN, where the scarf stripes are over enhanced. 

In Table \ref{tab:performance} we show the performance of our network as compared to the alternatives. Here, we have used the average per-image peak signal-to-noise ratio (PSNR) \cite{salomon2004data} and the structural similarity index (SSIM) \cite{wang2004image}. We have chosen these two image quality metrics due to a couple of reasons. Firstly, these have been used extensively for the evaluation of super-resolution results elsewhere in the literature. Secondly, the PSNR is a signal processing approach based upon the mean-squared error whereas the SSIM is a structural similarity measure. From the table, we can observe that despite LapSRN  is the best performer, our method is often no more than 2 decibels below LapSRN in terms of the PSNR and within a 0.05 difference in the SSIM.

Further, in %Table \ref{tab:timeplot} 
Figure \ref{fig:timeplot}, we show the average per-image testing time, in milliseconds, for the three test datasets under consideration and three upscale factors, \textit{i.e.} $2$, $3$ and $4$. For all testing datasets our network far outperforms the alternatives, being approximately an order of magnitude faster than LapSRN  and more than two orders of magnitude faster than SRCNN.

% \begin{table*}
% \begin{center}
% \begin{tabular}{|l|c|c|c|c|c|c|}
% \hline
%   & A+ & SRCNN & VDSR & LapSRN & Ours\\
% Dataset  & time & time & time  & time  &   time  \\
% \hline\hline
% Set5 	 &  0.38 &  2.21 &  0.13 &  0.08 &  0.009\\
% \hline
% Set14 	 &  0.60  &  4.37 &  0.25 &  0.11 &  0.012\\
% \hline
% B100    &  0.39  &  2.53  &  0.19 &  0.06 &  0.006\\
% \hline
% \end{tabular}
% \end{center}
% \caption{Time.   }
% \end{table*}

% \begin{table}
% \begin{center}
% \begin{tabular}{|l|c|c|c|c|c|c|}
% \hline
%   &  & A+ & SRCNN & VDSR & LapSRN & Ours\\
% Dataset & Scale & time & time & time  & time  &   time  \\
% \hline\hline
%   		& 2	 &  0.58 &  2.19 &  0.13 &  0.09 &  0.009   \\
% Set5 	     & 3 &  0.32 &  2.23 &  0.13 &  0.08 &  0.009\\
%   		& 4  &  0.24 &  2.19 &  0.12 &  0.08 &  0.009\\
% \hline\hline
%   		& 2 &  0.86 &  4.32 &  0.25 &  0.10 &  0.012\\
% Set14 	& 3  &  0.56  &  4.40 &  0.26 &  0.11 &  0.012\\
%   		& 4  &  0.38 &  4.39  &  0.25 &  0.11  &  0.012\\
% \hline\hline
%   		& 2  &  0.59  &  2.51 &  0.16 &   0.06 &  0.006\\
% B100      & 3  &  0.33  &  2.58  &  0.21 &  0.06 &  0.006\\
%   		& 4  & 0.26   &  2.51  &  0.21 &  0.06 &  0.006  \\
% \hline
% \end{tabular}
% \end{center}
% \caption{Time.   }
% \label{tab:timeplot}
% \end{table}

\section{Conclusions}

In this paper, we have presented a computationally efficient frequency domain neural network for super-resolution. The network can be viewed as a frequency domain analogue of spatial domain CNNs. To our knowledge, this is the first network of its kind, where rectifier units in the spatial domain are substituted by convolutions in the frequency domain and vice versa. Moreover, the network is quite general in nature and well suited for other applications in computer vision and image processing which are traditionally tackled in the frequency domain. We have presented results an comparison with alternatives elsewhere in the literature. In our experiments, our network is up to more than two orders of magnitude faster than the alternatives with an imperceptible loss of performance.

{\small
\bibliographystyle{ieee}
\bibliography{references}

\begin{thebibliography}{10}\itemsep=-1pt

\bibitem{baker:2002}
S.~Baker and T.~Kanade.
\newblock Limits on super-resolution and how to break them.
\newblock {\em IEEE Transactions on Pattern Analysis and Machine Intelligence},
  24(9):1167--1183, 2002.

\bibitem{bevilacqua2012low}
M.~Bevilacqua, A.~Roumy, C.~Guillemot, and M.~L. Alberi-Morel.
\newblock Low-complexity single-image super-resolution based on nonnegative
  neighbor embedding.
\newblock 2012.

\bibitem{bishop:09}
T.~Bishop, S.~Zanetti, and P.~Favaro.
\newblock Light field superresolution.
\newblock In {\em IEEE International Conference on Computational Photography},
  2009.

\bibitem{bose:93}
N.~K. Bose, H.~C. Kim, and H.~M. Valenzuela.
\newblock Recursive implementation of total least squares algorithm for image
  reconstruction from noisy, undersampled multiframes.
\newblock In {\em IEEE Conference on Acoustics, Speech and Signal Processing},
  volume~5, pages 269--272, 1993.

\bibitem{bracewell:84}
R.~N. Bracewell.
\newblock The fast hartley transform.
\newblock {\em Proceedings of the IEEE}, 72(9):1010–1018, 1984.

\bibitem{dong2016image}
C.~Dong, C.~C. Loy, K.~He, and X.~Tang.
\newblock Image super-resolution using deep convolutional networks.
\newblock {\em IEEE transactions on pattern analysis and machine intelligence},
  38(2):295--307, 2016.

\bibitem{eren:97}
P.~E. Eren, M.~I. Sezan, and A.~M. Tekalp.
\newblock Robust, object-based high resolution image reconstruction from
  low-resolution video.
\newblock {\em IEEE Transactions on Image Processing}, 6(10):1446--1451, 1997.

\bibitem{Everingham10}
M.~Everingham, L.~Van~Gool, C.~K.~I. Williams, J.~Winn, and A.~Zisserman.
\newblock The pascal visual object classes (voc) challenge.
\newblock {\em International Journal of Computer Vision}, 88(2):303--338, June
  2010.

\bibitem{farsiu:2003}
S.~Farsiu, D.~Robinson, M.~Elad, and P.~Milanfar.
\newblock Fast and robust multi-frame super-resolution.
\newblock {\em IEEE Transactions on Image Processing}, 13:1327--1344, 2003.

\bibitem{glorot2011deep}
X.~Glorot, A.~Bordes, and Y.~Bengio.
\newblock Deep sparse rectifier neural networks.
\newblock In {\em Proceedings of the Fourteenth International Conference on
  Artificial Intelligence and Statistics}, pages 315--323, 2011.

\bibitem{HartleyTrans42}
R.~Hartley.
\newblock A more symmetrical fourier analysis applied to transmission problems.
\newblock 30:144 -- 150, 04 1942.

\bibitem{huang2015single}
J.-B. Huang, A.~Singh, and N.~Ahuja.
\newblock Single image super-resolution from transformed self-exemplars.
\newblock In {\em Proceedings of the IEEE Conference on Computer Vision and
  Pattern Recognition}, pages 5197--5206, 2015.

\bibitem{kappeler:2016}
A.~Kappeler, S.~Yoo, Q.~Dai, and A.~K. Katsaggelos.
\newblock Video super-resolution with convolutional neural networks.
\newblock {\em IEEE Transactions on Computational Imaging}, 2(2):109--122,
  2016.

\bibitem{kim2016accurate}
J.~Kim, J.~Kwon~Lee, and K.~Mu~Lee.
\newblock Accurate image super-resolution using very deep convolutional
  networks.
\newblock In {\em Proceedings of the IEEE Conference on Computer Vision and
  Pattern Recognition}, pages 1646--1654, 2016.

\bibitem{kim:90}
S.~P. Kim, N.~K. Bose, and H.~M. Valenzuela.
\newblock Recursive reconstruction of high resolution image from noisy
  undersampled multiframes.
\newblock {\em IEEE Transactions on Acoustics, Speech and Signal Processing},
  38(6):1013--1027, 1990.

\bibitem{ImageNetAlex2012}
A.~Krizhevsky, I.~Sutskever, and G.~E. Hinton.
\newblock Imagenet classification with deep convolutional neural networks.
\newblock In F.~Pereira, C.~J.~C. Burges, L.~Bottou, and K.~Q. Weinberger,
  editors, {\em Advances in Neural Information Processing Systems 25}, pages
  1097--1105. Curran Associates, Inc., 2012.

\bibitem{LapSRN}
W.-S. Lai, J.-B. Huang, N.~Ahuja, and M.-H. Yang.
\newblock Deep laplacian pyramid networks for fast and accurate
  super-resolution.
\newblock In {\em IEEE Conference on Computer Vision and Pattern Recognition},
  2017.

\bibitem{lecun1998gradient}
Y.~LeCun, L.~Bottou, Y.~Bengio, and P.~Haffner.
\newblock Gradient-based learning applied to document recognition.
\newblock {\em Proceedings of the IEEE}, 86(11):2278--2324, 1998.

\bibitem{MartinFTM01}
D.~Martin, C.~Fowlkes, D.~Tal, and J.~Malik.
\newblock A database of human segmented natural images and its application to
  evaluating segmentation algorithms and measuring ecological statistics.
\newblock In {\em Proc. 8th Int'l Conf. Computer Vision}, volume~2, pages
  416--423, July 2001.

\bibitem{pascanu2013difficulty}
R.~Pascanu, T.~Mikolov, and Y.~Bengio.
\newblock On the difficulty of training recurrent neural networks.
\newblock In {\em International Conference on Machine Learning}, pages
  1310--1318, 2013.

\bibitem{poynton2012digital}
C.~Poynton.
\newblock {\em Digital video and HD: Algorithms and Interfaces}.
\newblock Elsevier, 2012.

\bibitem{protter:2009}
M.~Protter and M.~Elad.
\newblock Super resolution with probabilistic motion estimation.
\newblock {\em IEEE Transactions on Image Processing}, 18(8):1899--1904, 2009.

\bibitem{rippel2015spectral}
O.~Rippel, J.~Snoek, and R.~P. Adams.
\newblock Spectral representations for convolutional neural networks.
\newblock In {\em Advances in Neural Information Processing Systems}, pages
  2449--2457, 2015.

\bibitem{salomon2004data}
D.~Salomon.
\newblock {\em Data compression: the complete reference}.
\newblock Springer Science \& Business Media, 2004.

\bibitem{schulter2015fast}
S.~Schulter, C.~Leistner, and H.~Bischof.
\newblock Fast and accurate image upscaling with super-resolution forests.
\newblock In {\em Proceedings of the IEEE Conference on Computer Vision and
  Pattern Recognition}, pages 3791--3799, 2015.

\bibitem{simonyan:2014}
K.~Simonyan and A.~Zisserman.
\newblock Very deep convolutional networks for large-scale image recognition.
\newblock {\em CoRR}, 2014.

\bibitem{taiimage}
Y.~Tai, J.~Yang, and X.~Liu.
\newblock Image super-resolution via deep recursive residual network.

\bibitem{timofte2014a}
R.~Timofte, V.~De~Smet, and L.~Van~Gool.
\newblock A+: Adjusted anchored neighborhood regression for fast
  super-resolution.
\newblock In {\em Asian Conference on Computer Vision}, pages 111--126.
  Springer, 2014.

\bibitem{tsai:84}
R.~Y. Tsai and T.~S. Huang.
\newblock Multipleframe image restoration and registration.
\newblock In {\em Advances in Computer Vision and Image Processing}, pages
  317--339, 1984.

\bibitem{wang2004image}
Z.~Wang, A.~C. Bovik, H.~R. Sheikh, and E.~P. Simoncelli.
\newblock Image quality assessment: from error visibility to structural
  similarity.
\newblock {\em IEEE transactions on image processing}, 13(4):600--612, 2004.

\bibitem{yang2010image}
J.~Yang, J.~Wright, T.~S. Huang, and Y.~Ma.
\newblock Image super-resolution via sparse representation.
\newblock {\em IEEE transactions on image processing}, 19(11):2861--2873, 2010.

\bibitem{zeyde2010single}
R.~Zeyde, M.~Elad, and M.~Protter.
\newblock On single image scale-up using sparse-representations.
\newblock In {\em International conference on curves and surfaces}, pages
  711--730. Springer, 2010.

\end{thebibliography}
}

\end{document}